\pdfoutput=1

\documentclass[11pt]{article}
\usepackage{misc/ACL2023}
\usepackage[T2A,T1]{fontenc}
\usepackage[utf8]{inputenc}
\usepackage[russian,english]{babel}

\usepackage{times}
\usepackage{latexsym}
\usepackage{microtype}
\usepackage{inconsolata}
\usepackage[disable]{todonotes}
\usepackage{amsmath}
\usepackage{amssymb}
\usepackage{bbm} %
\usepackage{bm}
\usepackage{cleveref}
\usepackage{xcolor}
\usepackage{colortbl} %
\usepackage{booktabs}
\usepackage{lipsum}
\usepackage{enumitem}
\usepackage{xspace}
\usepackage{soul} %
\usepackage{caption}
\usepackage{transparent} %
\usepackage{tcolorbox} %
\usepackage{multirow}
\usepackage[normalem]{ulem}
\usepackage{mdframed} %
\usepackage{float}
\usepackage{array}
\usepackage{arydshln} %
\usepackage{refcount} %
\usepackage{tablefootnote} %

\usepackage{enumitem} %

\newcommand{\metrictype}[1]{\emph{#1}}

\definecolor{TodoColor}{rgb}{1,0.7,0.6}

\newmdenv[
  linecolor=black,
  linewidth=1.2pt,
  topline=false,
  bottomline=false,
  rightline=false,
  innertopmargin=0mm,
  innerbottommargin=-0.5mm,
  innerleftmargin=1mm,
  skipabove=1.1\topsep,
  skipbelow=0.5\topsep,
]{quotebox}

\newcommand{\bio}{bio\xspace}
\newcommand{\general}{WMT\xspace}

\definecolor{badred}{rgb}{0.9, 0.3, 0.3}
\definecolor{goodgreen}{rgb}{0.2, 0.6, 0.2}

\let\svthefootnote\thefootnote
\newcommand\blankfootnote[1]{%
  \let\thefootnote\relax\footnotetext{#1}%
  \let\thefootnote\svthefootnote%
}

\newcounter{footnotecounter}

\title{Fine-Tuned Machine Translation Metrics Struggle in Unseen Domains}

\author{\begin{tabular}{ccc} 
    Vilém Zouhar$^1$\thanks{\hspace{1.5mm}Work done during an internship at Amazon.} & Shuoyang Ding$^2$ & Anna Currey$^2$ \\
    Tatyana Badeka$^2$ & Jenyuan Wang$^2$ & Brian Thompson$^2\thanks{\hspace{1.5mm}Corresponding author}$
\end{tabular} \\
$^1$ETH Zürich\quad $^2$AWS AI Labs \\
\href{mailto:brianjt@amazon.com}{brianjt@amazon.com}}

\begin{document}
\maketitle

\begin{abstract}

We introduce a new, extensive multidimensional quality metrics (MQM) annotated dataset covering 11 language pairs in the biomedical domain. 
We use this dataset to investigate whether machine translation (MT) metrics which are fine-tuned on human-generated MT quality judgements %
are robust to domain shifts between training and inference. 
We find that fine-tuned metrics exhibit a substantial performance drop in the unseen domain %
scenario relative to both metrics that rely on the surface form and pre-trained metrics that are not fine-tuned on %
MT quality judgments.

\end{abstract}

\section{Introduction}
Automatic metrics are vital for machine translation (MT) research:\ given the cost and effort required for manual evaluation, automatic metrics are useful for model development and reproducible comparison between research papers~\citep{ma-etal-2019-results}. 
In recent years, the MT field has been moving away from string-matching metrics like \textsc{BLEU}~\citep{papineni-etal-2002-bleu} towards fine-tuned metrics like \textsc{Comet} \citep{rei-etal-2020-comet}, which start with pre-trained models and then fine-tune them on human-generated quality judgments. 
Fine-tuned metrics have been the best performers in recent WMT metrics shared task evaluations \cite{freitag-etal-2022-results, freitag-etal-2023-results} and 
are recommended by the shared task organizers, who go so far as to say, \textit{``Neural fine-tuned metrics are not only better, but also robust to different domains.''} (\citealp{freitag-etal-2022-results}).

Given the growing popularity of fine-tuned metrics, it is important to better understand their behavior. %
Here, we examine the question of domain robustness of fine-tuned metrics.
Fine-tuned metrics contain extra parameters on top of the pre-trained model which are initialized randomly (or to zero) and then fine-tuned on human-generated MT quality annotations.
The primary source of those annotations is prior WMT metrics shared tasks, and 
domains in WMT are often carried over from year to year (e.g.\ news). %
This raises the question: are fine-tuned metrics in fact robust across any domain (including domains not seen in training)? Or can their apparent strong performance 
be attributed in part to 
the artificially good domain match between training and test data?

\begin{figure}[t]
\centering
\vspace{-2mm}
\includegraphics[width=0.999\linewidth]{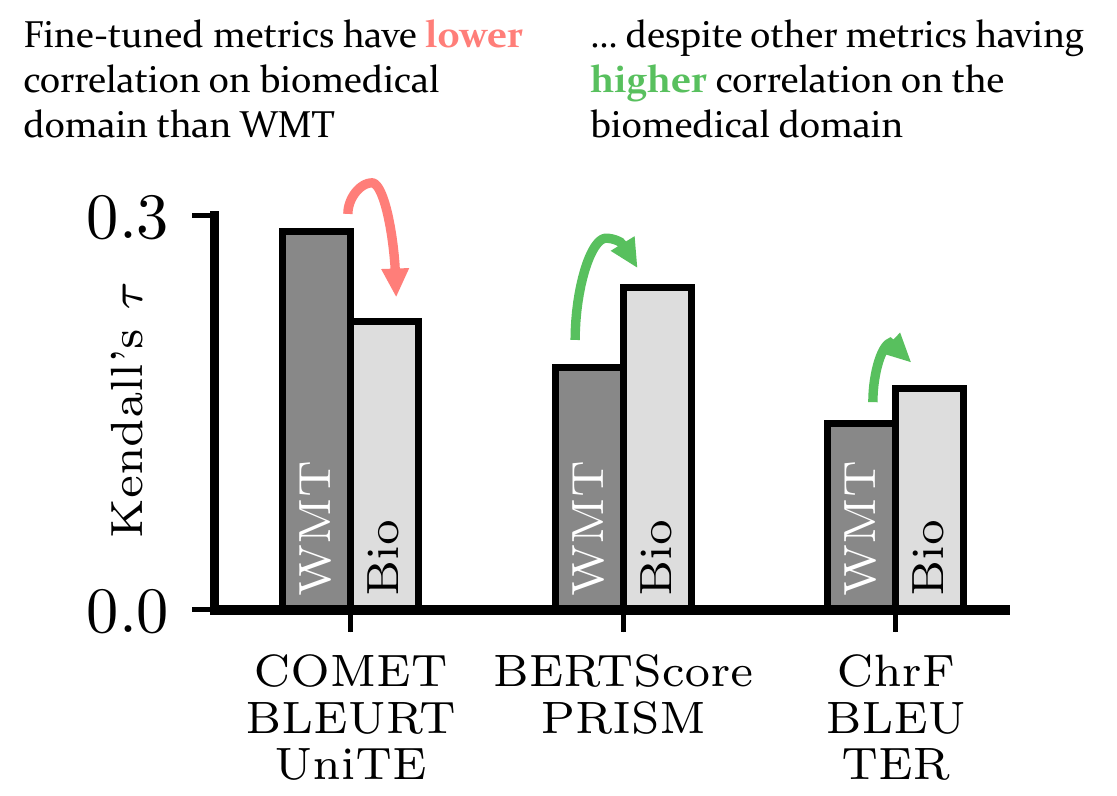}
\caption{Automatic machine translation metric performance on the WMT and biomedical domains, averaged across metric types (see \autoref{fig:domain_mismatch} for full results).}\label{fig:domain_mismatch_small}
\vspace{-3mm}
\end{figure}

To answer these questions, we first collect human multidimensional quality metrics (MQM) annotations in the biomedical (\bio) domain. Vocabulary overlap and error analysis suggest that this new dataset is distinct from the domains used in WMT.
This data covers 11 language pairs and 21 translation systems, with 25k total judgments. 
In addition to the MQM annotations, we also create new high-quality reference translations for all directions. 
We release this data publicly, along with code for replication of our experiments.\footnote{\href{https://github.com/amazon-science/bio-mqm-dataset}{\nolinkurl{github.com/amazon-science/bio-mqm-dataset}}}

Next, we examine how different types of metrics perform on our new \bio test set relative to the WMT test set. %
We find that fine-tuned metrics have substantially lower correlation with human judgments in the \bio domain, despite other types of metrics having higher correlation in the \bio domain 
(see \autoref{fig:domain_mismatch_small}), indicating they struggle with the training/inference domain mismatch.
Finally, we present analysis showing that this performance gap persists throughout different stages of the fine-tuning process and is not the result of a deficiency with the pre-trained model.

\newcommand{\smallemoji}[1]{\includegraphics[height=3mm]{img/emoji/#1.pdf}}

\begin{table} %
\centering
\resizebox{0.8\linewidth}{!}{\begin{tabular}{c<{\hspace{-1mm}} l}
\toprule
\bf Architecture & \bf Metrics \\
\midrule \\[-0.9em]
\shortstack[c]{Surface-Form\\\includegraphics[height=1.7em]{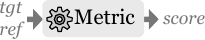}} &
\shortstack[l]{\textsc{BLEU}\\\textsc{ChrF}\\\textsc{TER}} \\[0.5em]

\shortstack[c]{Pre-trained+Algorithm\\\includegraphics[height=2.2em]{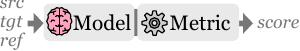}} &
\shortstack[l]{ \textsc{BERTScore} \\ \textsc{Prism} \\\,} \\[0.5em]

\shortstack[c]{Pre-trained+Fine-tuned\\\includegraphics[height=2.2em]{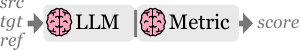}} &
\shortstack[l]{\textsc{Comet} \\ \textsc{UniTE} \\ \textsc{Bleurt}}  \\[0.5em]

\shortstack[c]{Pre-trained+Prompt\\\includegraphics[height=2.2em]{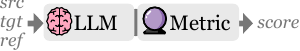}} &
\shortstack[l]{\textsc{GEMBA} \\ \textsc{AutoMQM} \\\,} \\

\bottomrule
\end{tabular}}
\caption{Metric types considered in this work. The \smallemoji{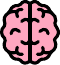} components have trainable parameters while \smallemoji{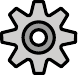} use handcrafted heuristics or algorithms and \smallemoji{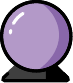} decodes from a language model.
The \textit{ref} input is omitted in the case of reference-free metrics (i.e.\ quality estimation).
}
\label{tab:metric_types}
\end{table}

\section{Related Work}

\paragraph{Metric types.}
\autoref{tab:metric_types} summarizes the different types of metrics that are commonly used to evaluate MT. 
The earliest type of MT metrics are \metrictype{Surface-Form} metrics, which are purely heuristic and use word- or character-based features.
We consider three common \metrictype{Surface-Form} metrics: \textsc{BLEU} \citep{papineni-etal-2002-bleu}, \textsc{TER} \citep{snover-etal-2006-study} and \textsc{ChrF} \citep{popovic-2015-chrf}.
Metrics like \textsc{Comet} \citep{rei-etal-2020-comet}, 
\textsc{Bleurt} \citep{sellam-etal-2020-bleurt}, 
and \textsc{UniTE} \citep{wan-etal-2022-unite} 
start with a pre-trained language model and fine-tune it on human-generated MT quality judgments. We denote these metrics \metrictype{Pre-trained+Fine-tuned}.\footnote{The WMT metrics task calls these ``trained'' metrics.}
Another class of metrics also start with a pre-trained model but do not perform fine-tuning. Examples of such metrics include \textsc{Prism} \citep{thompson-post-2020-automatic, thompson-post-2020-paraphrase}, which uses the perplexity of a neural paraphraser, and \textsc{BERTScore} \citep{sun-etal-2022-bertscore}, which is based on cosine similarity of word embeddings.
We denote such metrics \metrictype{Pre-trained+Algorithm} metrics.
More recently, metrics like \textsc{GEMBA} \citep{kocmi2023large} and \textsc{AutoMQM} \citep{fernandes2023devil} have proposed prompting a large language model. We denote these as \metrictype{Pre-trained+Prompt} metrics.

\paragraph{Domain specificity.}

Domain specificity for MT metrics was first explored by \citet{c-de-souza-etal-2014-machine} for \metrictype{Surface-Form} metrics.
\citet{sharami2023tailoring} brought attention to the issue of domain adaptation for quality estimation (QE), offering solutions based on curriculum learning and generating synthetic scores similar to \citet{heo-etal-2021-quality}, \citet{baek-etal-2020-patquest}, and \citet{zouhar-etal-2023-poor}.
\citet{sun-etal-2022-bertscore} examined general-purpose natural language generation metrics and documented their bias with respect to social fairness.
For word-level QE, \citet{sharami2023tailoring} reported the lack of robustness of neural metrics.

\section{New Bio MQM Dataset}\label{sec:biomedical_data}

We create and release new translations and MQM annotations for the system submissions from 21 participants to the WMT21 biomedical translation shared task \citep{yeganova-etal-2021-findings}. 
To explore how different the \bio domain is from the WMT22 metric task domains, we computed the vocabulary overlap coefficient between each domain. Bio had the smallest average overlap with the WMT domains (0.436) compared to 0.507, 0.486, 0.507, and 0.582 for e-commerce, news, social, and conversation, respectively. See \autoref{appendix:vocab} for full details and example sentences from each domain.

\begin{figure*} %
\centering
\includegraphics[width=0.9\textwidth]{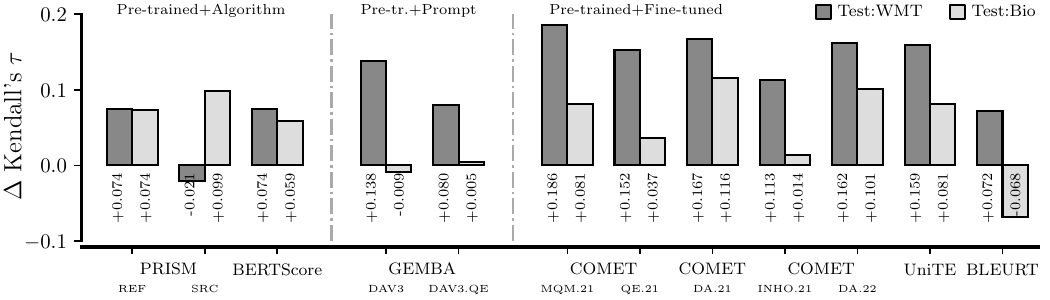}
\caption{Gains in segment-level correlation (Kendall's $\tau$) when comparing \metrictype{Surface-Form} metrics (average performance of \textsc{BLEU}, \textsc{ChrF}, and \textsc{TER}) to a given metric, on the \general and \bio test sets. 
Gains for \metrictype{Pre-trained+Fine-tuned} metrics are much smaller in  the unseen \bio domain than the \general domain. 
\metrictype{Pre-trained+Algorithm} metrics, which do not train on prior WMT data, do not exhibit the same bias. 
See \autoref{sec:raw_kendal_tao} for results in tabular form. 
}
\label{fig:domain_mismatch}
\end{figure*}

\subsection{Dataset Creation}

\begin{table}[t] %
\centering
\small
\begin{tabular}{
l>{\hspace{-2mm}}
l
>{\hspace{-2mm}}c
c
}
\toprule

&& \bf WMT & \bf Bio \\
\midrule
\multirow{4}{*}{\shortstack[l]{Error\\ severity}}
& Critical & N/A & 8\% \\
& Major & 26\% & 44\% \\
& Minor & 43\% & 31\% \\
& Neutral & 31\% & 16\% \\[0.7em]
\multirow{5}{*}{\shortstack[l]{Error\\ category}}
& Fluency & 47\% & 66\%  \\
& Accuracy & 44\% & 18\%  \\
& Terminology & 6\% & 10\%  \\
& Locale & 2\% & 2\%  \\
& Other & 1\% & 4\% \\[0.7em]
\multicolumn{2}{l}{Error-free segments} & 45\%  & 72\%\\
\multicolumn{2}{l}{Errors per erroneous segment} & 1.9 & 2.1\\
\multicolumn{2}{l}{Abs. erroneous segment score} &  -4.1\, & -7.6\,\\
\bottomrule
\end{tabular}
\caption{Error distribution of our new bio dataset and the existing WMT22 MQM dataset. The MQM annotation scheme for WMT in most cases did not contain the \textit{Critical} category.}
\label{tab:data_mqm_properties}
\end{table}

We created the bio MQM dataset in three steps.
Annotations and translations were performed by expert linguists with experience in the medical domain (see \autoref{appendix:qualifications} for full details). 

\paragraph{Step 1: Reference re-translation.}
The original \bio test set consists of bilingual abstracts from crawled academic papers, which might be written by non-native speakers \citep{neveol-etal-2020-medline} or even MT \cite{thompson2024shocking}.
Therefore, we create new professional reference translations. %

\paragraph{Step 2: Reference quality.}
To ensure a high bar of quality for the reference translations, we ask a separate set of annotators to provide MQM annotations for the new references. 
Any issues identified by this round of MQM annotation are then fixed by a new set of translators, resulting in the final reference translations that we release in this dataset. %

\paragraph{Step 3: MQM annotations.}
Finally, we conduct the main MQM annotation on the references and shared task system outputs.
In this step, a single annotator rates all translations of a given document (from all systems and the reference).\footnote{This allows us to distribute annotation jobs to multiple annotators while still allowing the annotator to access document-level context and ensuring that the whole document is ranked consistently.}
Our MQM schema follows \citet{freitag-etal-2021-experts} except that we add a \textit{Critical} severity (assigned the same score as \textit{Major} for backward compatibility). Full annotator instructions are in \autoref{appendix:guidelines}.

The resulting dataset contains roughly 25k segment-level annotations spanning 11 translation directions.\footnote{Pt$\rightarrow$En, En$\leftrightarrow$De, En$\leftrightarrow$Es, En$\leftrightarrow$Ru, En$\leftrightarrow$Fr, Zh$\leftrightarrow$En}
In contrast, most publicly available MQM data to date covers only a few language pairs. 
We use \textasciitilde25\% of the segments for each language pair as the train/dev set, leaving the rest as the test set (see \autoref{appendix:stats} for exact sizes in each pair).

We compare error distributions on our new bio MQM dataset and the existing WMT MQM dataset in \autoref{tab:data_mqm_properties}.
Bio MQM contains more \textit{Critical}/\textit{Major} errors, and lower absolute scores on average. However, WMT MQM has more overall sentences where an error occurs.
Error category distribution also diverges, notably in \textit{Fluency} and \textit{Accuracy}.

\section{Analysis}

\subsection{Are fine-tuned metrics robust across domains?} \label{sec:is-it-robust}

\paragraph{Measuring domain robustness.} The performance of a MT metric is typically measured by a certain \emph{meta-evaluation metric}, such as segment-level Kendall's $\tau$ correlation with human judgments. Intuitively, one could simply measure domain robustness by comparing the performance of a certain metric on domain A and domain B.
This, however, is not straightforward with meta-evaluations for metrics, since performance measured by those meta-evaluations is also affected by factors such as the quantity and quality of the translations included in the dataset, which is often hard to control for.

As a result, we resort to comparisons of \emph{relative} performance measured against a domain-invariant baseline. To establish such comparison, we make two assumptions:
\begin{enumerate}
    \item We assume \metrictype{Surface-Form} metrics can serve as a domain-invariant baseline, as they are purely based on heuristics and do not involve parameters specifically tuned on a certain domain. We use average performance of \textsc{BLEU}, \textsc{ChrF}, and \textsc{TER} as the baseline to minimize the impact of specific choice of heuristics.
    \item We assume segment-level Kendall's $\tau$ correlation with human judgments has a linear relationship with the objective performance of a metric. Hence, relative performance can be measured by simple linear subtraction.
\end{enumerate}

\paragraph{Observations.} Compared to \metrictype{Surface-Form} metrics, we find that \metrictype{Pre-trained+Fine-tuned} metrics provide 
a substantially smaller (sometimes even negative) improvement in human correlation in the \bio domain than the \general domain (see \autoref{fig:domain_mismatch}).
On the other hand, \metrictype{Pre-trained+Algorithm} metrics, which have not been trained on \general data, 
do not exhibit the same gap. This gap suggests that fine-tuned metrics struggle with unseen domains.

We also observe a very large performance gap for \metrictype{Pre-trained+Prompt} metrics.
Unfortunately, these metrics rely on closed-source LLMs without published training procedures, so we do not know what data the underlying LLMs were trained on. 

\subsection{How does fine-tuning affect domain robustness?}

\paragraph{Model description.} For this section, we focus on \textsc{Comet} (reference-based) and \textsc{Comet-QE} (reference-free) as they are among the most commonly used MT metrics.
The \textsc{Comet} model works by representing the source, the hypothesis and the reference as three fixed-width vectors using a language model, such as XLM-Roberta-large \cite{roberta}.
These vectors and their combinations serve as an input to a simple feed-forward regressor which is fine-tuned to minimize the MSE loss with human MQM scores.
A \textsc{Comet} model is trained in two stages, first on direct assessment (DA) quality annotations and then on MQM annotations, both from WMT shared tasks.

\paragraph{Setup.} We limit our experiments to the En-De, Zh-En and Ru-En language directions because of WMT MQM availability. We largely followed the training recipe in the \textsc{Comet} Github repo\footnote{\href{https://github.com/Unbabel/COMET/tree/master/configs}{\nolinkurl{github.com/Unbabel/COMET/tree/master/configs}}}. For details, please refer to our code.

There is high inter-annotator variance in the WMT and bio MQM data.
Training on the raw MQM scores is very unstable and therefore per-annotator z-normalizing is necessary to replicate our setup.
Note that the publicly available WMT MQM data are not z-normalized.

\paragraph{Observation 1: Domain gap persists throughout the fine-tuning process.} \label{sec:fine-tuning}

\newcommand{\zerowidthsymbol}[1]{\hspace{0.5mm}\makebox[0pt]{$\boldsymbol#1$}\hspace{-1mm}\null}
\begin{table} %
\centering
\resizebox{0.9\linewidth}{!}{
\renewcommand*{\arraystretch}{1.1}
\setlength\tabcolsep{1.2mm}
\setlength{\arrayrulewidth}{.1em}
\begin{tabular}{
cc|
c<{\hspace{1.5mm}}
c<{\hspace{1.5mm}}
c<{\hspace{1.5mm}}
c<{\hspace{1.5mm}}
c<{\hspace{1.5mm}}
c<{\hspace{1.5mm}}
}

\textbf{\large Test:WMT}\hspace{-2cm} & \multicolumn{6}{c}{MQM epochs} \\
\parbox[t]{2mm}{\multirow{6}{*}{\rotatebox[origin=c]{90}{DA epochs\hspace{5mm}}}}
 &  & 0 & 1 & 2 & 4 & 8\\
\hline
&0 & 
\cellcolor{black!0.00} 0.118 & \cellcolor{black!4.90} 0.285 & \cellcolor{black!2.06} 0.281 & \cellcolor{black!0.00} 0.279 & \cellcolor{black!14.53} 0.295\\
&1 & 
\cellcolor{black!39.79} 0.324 & \cellcolor{black!48.10} 0.333 & \cellcolor{black!35.06} 0.318 & \cellcolor{black!33.33} 0.317 & \cellcolor{black!39.42} 0.323\\
&2 & 
\cellcolor{black!41.96} 0.326 & \cellcolor{black!51.52} 0.337 & \cellcolor{black!39.16} 0.323 & \cellcolor{black!39.23} 0.323 & \cellcolor{black!41.17} 0.325\\
&4 & 
\cellcolor{black!37.87} 0.322 & \cellcolor{black!50.00} 0.335 & \cellcolor{black!38.68} 0.323 & \cellcolor{black!38.14} 0.322 & \cellcolor{black!36.97} 0.321\\
&8 & 
\cellcolor{black!28.08} 0.311 & \cellcolor{black!49.87} 0.335 & \cellcolor{black!40.22} 0.324 & \cellcolor{black!38.04} 0.322 & \cellcolor{black!32.57} 0.316 \\
\\[-0.5em]
\textbf{\large Test:Bio}\hspace{-1.5cm} & \multicolumn{6}{c}{MQM epochs} \\
\parbox[t]{2mm}{\multirow{6}{*}{\rotatebox[origin=c]{90}{DA epochs\hspace{5mm}}}}
 &  & 0 & 1 & 2 & 4 & 8\\
\hline
&0 & 
\cellcolor{black!0.00} 0.071 & \cellcolor{black!4.66} 0.234 & \cellcolor{black!0.00} 0.229 & \cellcolor{black!10.64} 0.240 & \cellcolor{black!20.11} 0.250\\
&1 & 
\cellcolor{black!50.47} 0.282 & \cellcolor{black!48.02} 0.280 & \cellcolor{black!50.00} 0.282 & \cellcolor{black!42.90} 0.274 & \cellcolor{black!38.34} 0.270\\
&2 & 
\cellcolor{black!38.49} 0.270 & \cellcolor{black!33.92} 0.265 & \cellcolor{black!41.49} 0.273 & \cellcolor{black!37.31} 0.268 & \cellcolor{black!34.66} 0.266\\
&4 & 
\cellcolor{black!24.14} 0.255 & \cellcolor{black!16.37} 0.246 & \cellcolor{black!27.31} 0.258 & \cellcolor{black!28.19} 0.259 & \cellcolor{black!22.87} 0.253\\
&8 & 
\cellcolor{black!10.22} 0.240 & \cellcolor{black!12.16} 0.242 & \cellcolor{black!30.10} 0.261 & \cellcolor{black!29.50} 0.260 & \cellcolor{black!23.00} 0.253 

\end{tabular}
}
\caption{Segment-level correlation (Kendall's $\tau$) between metrics and human judgments on the \general (top) and \bio (bottom) test sets, for \textsc{Comet} with varying epochs of \general domain DA and MQM training.}
\label{tab:grid_dax_mqmx}
\end{table}

We would like to understand which stage among the two training stages for \textsc{Comet} accounts for the domain gap.
To this end, we retrained \textsc{Comet} with varying epochs on DA/MQM data, shown in Table \ref{tab:grid_dax_mqmx}.
In contrast to catastrophic forgetting \cite{goodfellow2013empirical, thompson-etal-2019-overcoming, thompson-etal-2019-hablex}, where a model starts with good general-domain performance and then overfits while being adapted to a new task or domain, we do not see a sharp dropoff in the bio domain performance when training on more \general (DA and/or MQM) data. This indicates that the model is a weak \bio metric at all stages, as opposed to first learning and then forgetting.

\paragraph{Observation 2: In-domain data dramatically improves \textsc{Comet}.}

\begin{figure} %
\includegraphics[width=0.99\linewidth]{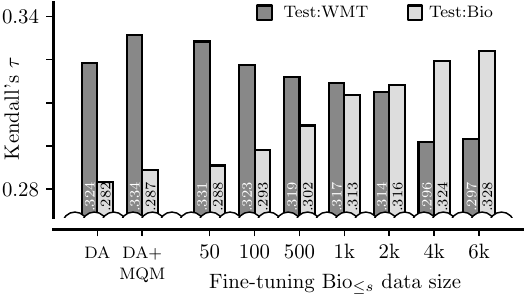}
\caption{Average performance (8 seeds) of \textsc{Comet} fine-tuned on varying amounts of MQM \bio data. }
\label{fig:adapt_mqm_finetune_datasize}
\end{figure}

Generally, including \bio MQM annotations in training improves \textsc{Comet}'s performance in the \bio test set, increasing correlation from 0.287 to 0.328 with 6k \bio judgments. Indeed, just 1k judgements improves correlation to 0.313 (see \autoref{fig:adapt_mqm_finetune_datasize}). 
This rules out the possibility that \bio is inherently problematic for \textsc{Comet}'s architecture or fine-tuning strategy.

\subsection{How does the pre-trained model affect domain robustness?}

\textsc{Comet} and \textsc{BERTScore} are both based on XLM-Roberta-large \cite{roberta}, allowing us to explore how the same changes to the pre-trained model affect each metric. To see whether improving the underlying pre-trained model improves \metrictype{Pre-trained+Algorithm} metrics built on those pre-trained models, we fine-tune XLM-Roberta with data similar to the WMT and \bio domain setup, respectively.
Similarly, we also investigate how \textsc{Prism}, another \metrictype{Pre-trained+Algorithm} metric,
is affected with changes to the pre-trained model. We use \textsc{Prism} with the NLLB multilingual MT models \citep{nllbteam2022language} as they are larger and more recent than the model released with \textsc{Prism}. 

\paragraph{Setup.} Our fine-tuning data covers the four languages of interest, namely English, German, Russian, and Chinese (see Appendix \ref{appendix:improve_bert} for a detailed data list).
Since NLLB is a translation model, we use only parallel data to fine-tune the model.
For the XLM-Roberta case, note that it was fine-tuned with two objectives: masked language model (MLM) and translation language model (TLM).
We use both parallel and monolingual data for MLM training and parallel data for TLM training.

\begin{figure} %
\includegraphics[width=0.99\linewidth]{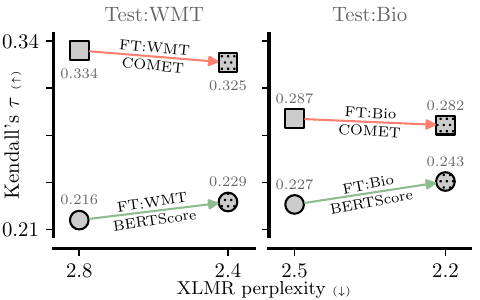}
\caption{Metric performance when pre-trained model is fine-tuned (FT) on bio or WMT domain data. 
\textcolor{goodgreen}{Lower} perplexity \textcolor{goodgreen}{improves} \textsc{BERTScore} $\bigcirc$ but \textcolor{badred}{worsens} \textsc{Comet} $\square$. Perplexity is average of MLM and TLM objectives on the text portion of the MQM dataset for both domains.}
\label{fig:bertscore_ppl}
\end{figure}

\begin{figure} %
\centering
\includegraphics[width=\linewidth]{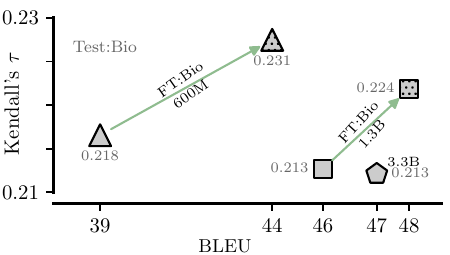}
\caption{Multiple NLLB MT models are used as the base model for \textsc{PrismSrc}. Fine-tuning the underlying MT model \textcolor{goodgreen}{improves the metric}. Compute constraints preclude finetuning NLLB-3.3B.}
\label{fig:prism_mts}
\end{figure}

\paragraph {Observations: XLM-Roberta.} For both domains, improving the pre-trained model 
improves 
\textsc{BERTScore}
but not 
\textsc{Comet} (see \autoref{fig:bertscore_ppl}).
This indicates that the limiting factor for the poor performance of \textsc{Comet} on \bio is the effect from its various fine-tuning stages (discussed in Section \ref{sec:fine-tuning}), not an underlying weakness in the pre-trained model on \bio.

\paragraph{Observations: NLLB.} Our findings are shown in \autoref{fig:prism_mts}. In general, we found that improving the pre-trained models performance (as measured by \textsc{BLEU} on a held out test set) also improved \textsc{Prism}'s performance.

\section{Conclusion and Future Work}
This paper investigated the performance of machine translation metrics across divergent domains. 
To this end, we introduced a new, extensive MQM-annotated dataset covering 11 language pairs in the \bio domain. Our analysis showed that \metrictype{Pre-trained+Fine-tuned} metrics (i.e.\ those that use prior human quality annotations of MT output) exhibit a larger gap between in-domain and out-of-domain performance than \metrictype{Pre-trained+Algorithm} metrics (like \textsc{BERTScore}). 
Further experiments showed that this gap can be attributed to the DA and MQM fine-tuning stage.

Despite the gap between in-domain and out-of-domain performance, \textsc{Comet} is still the best performing metric on the \bio domain in absolute terms. 
Thus, our findings suggest potential directions for future work including collecting more diverse human judgments for \metrictype{Pre-trained+Fine-tuned} metrics and exploring ways to improve the generalization of such metrics during fine-tuning. %

\section*{Limitations}

Our findings are dependent on two empirical assumptions we discussed in section \ref{sec:is-it-robust}. To the best of our knowledge, those assumptions are necessary to achieve a fair comparison of metrics across domains, but conclusions may change if our assumptions are refuted in future studies.

We draw conclusions based on a single unseen domain (biomedical). While additional domains would have been preferable, data collection was cost prohibitive. 

Context has been shown to be beneficial in machine translation evaluation \cite{laubli-etal-2018-machine, toral-2020-reassessing} and some metrics used in this work have document-level versions
\cite{vernikos-etal-2022-embarrassingly}.
However, in order to draw fair comparisons with existing metrics which do not yet have a document-level version, we only evaluated metrics at the sentence level.

We focused on segment-level evaluation and did not attempt system-level comparisons because of 
the limited number of system submissions to the WMT biomedical translation shared task.

\section*{Acknowledgements}

We would like to thank Georgiana Dinu, Marcello Federico, Prashant Mathur, Stefano Soatto, and other colleagues for their feedback at different stages of drafting.

\section*{Ethical Considerations}

Our human annotations were conducted through a vendor. Annotators were compensated in accordance to the industry standard -- specifically, in the range of \$27.50 to \$37.50 on an hourly basis, depending on the experience of the annotator.

\bibliography{main.bbl}
\bibliographystyle{misc/acl_natbib}

\clearpage
\appendix

\section{Domain Overlap Between WMT and \bio}\label{appendix:vocab}

To evaluate the overlap between the WMT and \bio domains, we calculate the vocabulary overlap coefficient $(\frac{|A \cap B|}{\min(|A|, |B|)})$ between our new bio MQM dataset and the domains used in the WMT22 metrics shared task. The per-domain overlap matrix is shown in \autoref{fig:vocab}. 
Randomly selected sentences from each domain are provided for illustration in \autoref{fig:sents}.

\begin{figure*}
\centering
\begin{tabular}{lrrrrr}
\toprule
 & \bf{e-commerce} & \bf{news} & \bf{social} & \bf{conversation} & \bf{biomedical} \\
\midrule
\bf{e-commerce} & 1.000 & 0.349 & 0.511 & 0.662 & 0.369 \\
\bf{news} & 0.349 & 1.000 & 0.517 & 0.592 & 0.359 \\
\bf{social} & 0.511 & 0.517 & 1.000 & 0.494 & 0.462 \\
\bf{conversation} & 0.662 & 0.592 & 0.494 & 1.000 & 0.554 \\
\bf{biomedical} & 0.369 & 0.359 & 0.462 & 0.554 & 1.000 \\
\bottomrule
\end{tabular}\caption{Vocabulary overlap coefficient between the English source-side data for each domain in the WMT22 and our \bio dataset. }\label{fig:vocab}
\end{figure*}

\begin{figure*}
\centering
\begin{tabular}{p{1.0in}p{4.8in}}
\toprule
\bf{e-commerce} & {This was one of the first albums I purchased of Keith's "back in the day".}\\
\midrule
\bf{news} & {Sean Combs has been variously known as Puff Daddy, P. Diddy or Diddy, but this year announced his preference for the names Love and Brother Love.}\\
\midrule
\bf{social} & {The comment about boiling being inefficient is probably correct bc even though the water heater is running continuously, that thing has SO MUCH insulation.}\\
\midrule
\bf{conversation} & {Let me know if you were able to create your new password and sign in with it}\\
\midrule
\bf{biomedical} & {Though neither perfectly sensitive nor perfectly specific for trachoma, these signs have been essential tools for identifying populations that need interventions to eliminate trachoma as a public health problem.}\\
\bottomrule
\end{tabular}\caption{Randomly selected English example sentences from each domain in the WMT22 metrics shared task as well as our new \bio dataset.}\label{fig:sents}
\end{figure*}

\section{Corpus Statistics}\label{appendix:stats}
\autoref{tab:data_sizes} shows the size per language pair of our bio MQM dataset, as well as the WMT MQM dataset for comparison. The bio MQM dataset contains roughly 25k annotated segments, covering 11 language pairs. We split the data into test (roughly 75\%) and development (roughly 25\%) sets.

\begin{table}[t] %
    \centering
    \small
    \begin{tabular}{cccccccc}
    \toprule
    & \multicolumn{2}{c}{\bf WMT} & \multicolumn{3}{c}{\bf Bio} \\
    \bf Langs & \bf Test & \bf Train  & \bf Test & \bf Dev & \bf Total\\
    \midrule
    De-En & - & - & 2457 & 903 & 3360\\
    En-De & 18k & 28k & 2695 & 917 & 3612\\
    Es-En & - & - & 1013 & 309 & 1322\\
    En-Es & - & - & 1112 & 330 & 1442\\
    Ru-En & - & - & 1324 & 388 & 1712\\
    En-Ru & 19k & 16k & 825 & 237 & 1062\\
    Fr-En & - & - & 1108 & 352 & 1460\\
    En-Fr & - & - & 1228 & 308 & 1536\\
    Zh-En & 23k & 27k& 2838 & 913 & 3751\\
    En-Zh & - & - & 3900 & 1200 & 5100\\
    Pt-En & - & - & 701 & 222 & 924 \\
    \midrule
    All & 60k & 71k & 19k & 6k & 25k \\
    \bottomrule
    \end{tabular}
    \caption{Data split of the bio MQM data released in this work, and WMT22 MQM \citep{freitag-etal-2022-results} data. All test results are reported with the \textit{test} split which is approximately $75\%$ of \textit{total}. Splits were created to respect document-level boundaries. For WMT, 2022 is used for testing and 2020 and 2021 for training.}
    \label{tab:data_sizes}
    \vspace{2mm}
\end{table}

\section{Translator/Annotator Qualifications}\label{appendix:qualifications}

There were 2-4 MQM annotators for each language pair, and a total of 46 annotators. All linguists had experience in translating/post-editing/reviewing content in the \bio domain. This was the main requirement to be able to work on the project.
The other qualification criteria for this project were in line with the ISO standard 17100. In particular, the linguists met one or more of the following criteria:
(1) A recognized higher education degree in translation;
(2) Equivalent third-level degree in another subject plus a minimum of two years of documented professional translation experience;
(3) A minimum of five years of documented professional translation experience;
(4) Native speaker of the target language.
Although linguists were experts in the \bio domain, not all of them were experts in MQM annotation. For this reason, the annotators completed an MQM quiz before onboarding them to ensure they understood the guidelines and requirements.

For the translation and post-editing tasks, we used a two step process (initial post editor + reviewer). In each case the reviewer was a linguist with experience translating medical texts. There were no specific educational or vocational stipulations on that medical qualification, however they were asked to provide a medical-text-specific translation test for us to be onboarded for the project.
The initial post-editor in each case was a linguistic expert, but not specifically an expert in medical translations, which is why we followed up with reviewers to ensure contents were translated accurately.
Linguists had to demonstrate the following to onboard to the project:
(1) At least 3+ years of professional translation experience
(2) Proven proficiency in English writing skills
(3) In-depth understanding and exposure to the language
(4) Strong ability in translating, reviewing, adjusting, and providing adaptation for various writing styles of particular requests.

\section{MQM Annotation Guidelines}
\label{appendix:guidelines}

Below, we reproduce the MQM annotation guidelines that we provided to the annotators.

\paragraph{Overview:}
You are asked to evaluate the translations using the guidelines below, and assign error categories and severities considering the context segments available. 

\paragraph{Task:}
\begin{enumerate}
    \item Please identify all errors within each translated segment, \underline{up to a maximum of five}.
    \begin{enumerate}
        \item If there are more than five errors, identify only the five most severe.
        \item If it is not possible to reliably identify distinct errors because the translation is too badly garbled or is unrelated to the source, then mark a single Unintelligible error that spans the entire segment
        \item Annotate segments in natural order, as if you were reading the document. You may return to revise previous segments.
    \end{enumerate}
    \item To identify an error, highlight the relevant span of text.
    \begin{enumerate}
        \item Omission and Source error should be tagged in the source text.
        \begin{enumerate}
            \item All other errors should be tagged in the target text.
        \end{enumerate}
        \item Unintelligible error should have an entire sentence tagged; if you think a smaller span is needed, then you should select another error category (Mistranslation, etc.).
    \end{enumerate}
    \item Select a category/sub-category and severity level from the available options.
    \item When identifying errors, please be as fine-grained as possible. 
    \begin{enumerate}
        \item If a sentence contains more than one error of the same category, each one should be logged separately. For example, if a sentence contains two words that are each mistranslated, two separate mistranslation errors should be recorded. 
        \item If a single stretch of text contains multiple errors, you only need to indicate the one that is most severe.
        \begin{enumerate}
            \item If all have the same severity, choose the first matching category listed in the error typology (e.g.\ Accuracy, then Fluency, then Terminology, etc.). 
        \end{enumerate}
        \item For repetitive errors that appear systematically through the document: please annotate each instance with the appropriate weight.
    \end{enumerate}
    \item Please pay particular attention to the context when annotating.  You will be shown several context segments before and after the segment for evaluation. If a translation is questionable on its own but is fine in the context of the document, it should not be considered erroneous; conversely, if a translation might be acceptable in some context, but not within the current document, it should be marked as wrong.
\end{enumerate}

\paragraph{Delivery format:}
\begin{itemize}
    \item file format: a TSV with additional columns for error categories and severity + JSON
    \begin{itemize}
        \item for multiple errors in one segment:  additional row for each error + severity
        \item text spans  will be highlighted for the annotation process and exported as tag
    \end{itemize}
\end{itemize}

\paragraph{Error categories:} Table~\ref{tab:mqm_error_categories}

\begin{table*}
    \centering\small
    \begin{tabular}{p{4cm}|p{2.5cm}p{6cm}|p{2cm}}
        & & & Tag\\
        & & & Location \\\hline
        \multirow{4}{4cm}{\textbf{Accuracy} – errors occurring when the target text does not accurately correspond to the propositional content of the source text, introduced by distorting, omitting, or adding to the message}
        &\textbf{Mistranslation}&Target content that does not accurately represent the source content.&Target\\
        &\textbf{Addition}&Target content that includes content not present in the source.&Target\\
        &\textbf{Omission}&Errors where content is missing from the translation that is present in the source.&\textbf{\textit{Source}}\\
        &\textbf{Untranslated}&Errors occurring when a text segment that was intended for translation is left untranslated in the target content.&Target\\ \hline

        \multirow{5}{4cm}{\textbf{Linguistic Conventions (former Fluency)} - errors related to the linguistic well-formedness of the text, including problems with, for instance, grammaticality and mechanical correctness.}&\textbf{Grammar}
        &Error that occurs when a text string (sentence, phrase, other) in the translation violates the grammatical rules of the target language.&Target\\
        &\textbf{Punctuation}&Punctuation incorrect for the locale or style.&Target\\
        &\textbf{Spelling}&Error occurring when the letters in a word in an alphabetic language are not arranged in the normally specified order.&Target\\
        &\textbf{Character encoding}&Error occurring when characters garbled due to incorrect application of an encoding.&Target\\
        &\textbf{Register}&Errors occurring when a text uses a level of formality higher or lower than required by the specifications or by common language conventions.&Target\\ \hline

        \multirow{2}{4cm}{\textbf{Terminology} - errors arising when a term does not conform to normative domain or organizational terminology standards or when a term in the target text is not the correct, normative equivalent of the corresponding term in the source text.} &&&\\
        &\textbf{Inconsistent use of terminology}&Use of multiple terms for the same concept (technical terms, medical terms, etc.)&Target\\&&&\\&&&\\
        &\textbf{Wrong term}&Use of term that it is not the term a domain expert would use or because it gives rise to a conceptual mismatch.&Target\\ &&&\\\hline

        \textbf{Style}&
        \textbf{Non-fluent}&Text does not sound fluent or natural as if it were translated by a non-native speaker or because the translation is following the source too closely.&Target\\\hline

        \multirow{7}{4cm}{\textbf{Locale Conventions} - errors occurring when the translation product violates locale-specific content or formatting requirements for data elements.}
        &\textbf{Number format}&&Target\\
        &\textbf{Currency format}&&Target\\
        &\textbf{Measurement format}&&Target\\
        &\textbf{Time format}&&Target\\
        &\textbf{Date format}&&Target\\
        &\textbf{Address format}&&Target\\
        &\textbf{Telephone format}&&Target\\ \hline

        \textbf{Other} && any error that does not fit the categories above & Target \\ \hline

        \textbf{Source errors} & \textbf{source error} & The error that occurs in the source. All source errors (e.g.\ non-fluent source) should be annotated as source errors — no sub-categories need to be selected.
\textbf{If the source error caused a target error: }
- if the source error and target errors belong to the same category, then only flag the source. 
-If source and target errors belong to different categories - even if you know that the source error caused the translation error - do flag both. & \textbf{\textit{Source}}\\ \hline

        \textbf{Unintelligible } && So many errors, or errors are so outrageous, that text becomes incomprehensible, and it is hard to pinpoint a specific error type. & Target. Tag the entire sentence. If the span is smaller, then a different category should be applied, such as Mistranslation, Untranslated, etc.
    \end{tabular}
    \caption{MQM error categories provided in annotator instructions.}
    \label{tab:mqm_error_categories}
\end{table*}

\paragraph{Severity (no weights, just severity):}
Table~\ref{tab:mqm_severities}

\begin{table*}
    \centering\small
    \begin{tabular}{l|p{6cm}p{3cm}p{3cm}}
         \textbf{severity}& \textbf{Definition} & \textbf{Source example} & \textbf{Translation example} \\ \hline
         \textbf{Neutral} & Neutral issues are items that need to be noted for further attention or fixing but which should not count against the translation. This severity level can be perceived as a flag for attention that does not impose a penalty. 
It should be used for “preferential errors” (i.e, items that are not wrong, per se, but where the reviewer or requester would like to see a different solution). & 
Source: Join us in celebrating 10 years of the company! & 
Target: Join us to celebrate 10 years of the company! \\ \hline

    \textbf{Minor} & 
Minor issues are issues that do not impact usability or understandability of the content.  If the typical reader/user is able to correct the error reliably and it does not impact the usability of the content, it should be classified as minor. & 
 
S1: Accurately distinguish between legitimate and high\textbf{-r}isk account registrations

S2: See how organizations worldwide are using fraud detection.

 &  
T1: Accurately distinguish between legitimate and high\textbf{- r}isk account registrations

T2: See how organization worldwide are using fraud detection. \\ \hline

\textbf{Major} & errors that would impact usability or understandability of the content but which would not render it unusable. For example, a misspelled word that may require extra effort for the reader to understand the intended meaning but does not make it impossible to comprehend should be labeled as a major error. Additionally, if an error cannot be reliably corrected by the reader/user (e.g., the intended meaning is not clear) but it does not render the content unfit for purpose, it should be categorized as major. &  
 
Source: Set the performance to 50 percent 
  &  
 
Target: Set performance 50 percent  \\ \hline

\textbf{Critical} & errors that would render a text unusable, which is determined by considering the intended audience and specified purpose. For example, a particularly bad grammar error that changes the meaning of the text would be considered Critical. Critical errors could result in damage to people, equipment, or an organization’s reputation if not corrected before use. If the error causes the text to become unintelligible, it would be considered Critical. & S1: Set the device on the highest temperature setting. 

S2: The next step would be to identify the point of leakage. 

S3: 1.3 degrees & 
T1: Set the device on the lowest temperature setting. 

T2: It would be to identify the next point of leakage.

T3: 1,300 degrees
\\ 
 
    \end{tabular}
    \caption{Severity examples and explanations provided in MQM annotation instructions.}
    \label{tab:mqm_severities}
\end{table*}

\section{Supplementary Information on Experiments}

\subsection{Training Steps and Compute Time for Experiments}

The overall training consists of the following steps (compute times using a single A10 GPU). The times are per epoch and some experiments require training for multiple epochs.

\begin{itemize}[left=0mm,noitemsep]
\item Language modeling $\rightarrow$ XLM-Roberta, 10hr/ep.
\item DA scores regression $\rightarrow$ \textsc{CometDA}, 10hr/ep.
\item MQM scores regression $\rightarrow$ \textsc{Comet}, 1hr/ep.
\end{itemize}

\subsection{List of Data for Fine-Tuning Pre-Trained Model}\label{appendix:improve_bert}

For WMT domain, we used news-commentary v18.1 dataset\footnote{\href{https://data.statmt.org/news-commentary/v18.1/}{\nolinkurl{data.statmt.org/news-commentary/v18.1/}}} for all languages.
For the bio domain, we list the data in Table \ref{tab:bio-data-xlmr-finetune}.

\begin{table}[H]
\centering
\scalebox{0.77}{
\begin{tabular}{@{}llll@{}}
\toprule
Data Type                    & Language(s) & Dataset                                                                                                                 & Lines \\ \midrule
\multirow{4}{*}{Parallel}    & en-de       & \begin{tabular}[c]{@{}l@{}}UFAL Medical Corpus\\ \cite{yeganova-etal-2021-findings}\end{tabular}       & 3M    \\
                             & en-de       & \multirow{3}{*}{\begin{tabular}[c]{@{}l@{}}MEDLINE\\ \cite{yeganova-etal-2021-findings}\end{tabular}}  & 35k   \\
                             & en-ru       &                                                                                                                         & 29k   \\
                             & en-zh       &                                                                                                                         & 19k   \\ \midrule
\multirow{4}{*}{Monoling.} & En          & CORD \cite{wang-etal-2020-cord}                                                                        & 1M    \\
                             & De          & \begin{tabular}[c]{@{}l@{}}Animal Experiments\tablefootnote{\href{https://www.openagrar.de/receive/openagrar_mods_00046540?lang=en}{\nolinkurl{www.openagrar.de/receive/openagrar_mods_00046540?lang=en}}}\\ GERNERMED\\ \cite{info:doi/10.2196/39077}\end{tabular} & 250k  \\
                             & Ru          & Medical QA                                                                                                              & 250k  \\
                             & Zh          & Chinese Medical Dataset\tablefootnote{\href{https://huggingface.co/datasets/shibing624/medical}{\nolinkurl{huggingface.co/datasets/shibing624/medical}}}                         & 2M    \\ \bottomrule
\end{tabular}
}
\caption{Collection of \bio domain data used in pre-trained model fine-tuning experiments.} \label{tab:bio-data-xlmr-finetune}
\end{table}

\section{Raw Scores for \autoref{fig:domain_mismatch}}\label{sec:raw_kendal_tao}

The segment-level correlation (Kendall's $\tau$) scores used to compute improvements in \autoref{fig:domain_mismatch} are provided in \autoref{tab:raw_kendal_tao}. Note that there is no public \textsc{Comet} 22 MQM model.

\begin{table*}
\small
\centering
\begin{tabular}{llcc}
\toprule
\bf Type & \bf Metric & \bf Test:WMT & \bf Test:Bio \\
\midrule
\multirow{ 3 }{*}{Surface-Form}
 & BLEU & 0.134 & 0.213 \\
 & ChrF & 0.151 & 0.192 \\
 & TER & 0.140 & 0.100 \\
\cmidrule{2-4}\\[-1em]
\multirow{ 3 }{*}{Pre-trained+Algorithm}
 & PRISM{\tiny REF} & 0.216 & 0.242 \\
 & PRISM{\tiny SRC} & 0.121 & 0.267 \\
 & BERTScore & 0.216 & 0.227 \\
\cmidrule{2-4}\\[-1em]
\multirow{ 2 }{*}{Pre-trained+Prompt}
 & GEMBA{\tiny DAV3} & 0.280 & 0.159 \\
 & GEMBA{\tiny DAV3.QE} & 0.222 & 0.173 \\
\cmidrule{2-4}\\[-1em]
\multirow{ 7 }{*}{Pre-trained+Fine-tuned}
 & COMET{\tiny MQM.21} & 0.328 & 0.249 \\
 & COMET{\tiny QE.21} & 0.294 & 0.205 \\
 & COMET{\tiny DA.21} & 0.309 & 0.284 \\
 & COMET{\tiny INHO.21} & 0.255 & 0.182 \\
 & COMET{\tiny DA.22} & 0.304 & 0.269 \\
 & UniTE & 0.301 & 0.249 \\
 & BLEURT & 0.214 & 0.100 \\
\bottomrule
\end{tabular}
\caption{Segment-level correlation (Kendall's $\tau$) between metrics and human judgments on the \general and \bio domain.
\metrictype{Pre-trained+Fine-tuned} metrics have lower correlation on \bio than on \general, 
while \metrictype{Surface-Form} and \metrictype{Pre-trained+Algorithm} tend to 
have higher correlation. }
\label{tab:raw_kendal_tao}
\end{table*}

\end{document}